\documentclass[sigconf]{acmart}
\makeatletter
\def\@ACM@checkaffil{% Only warnings
    \if@ACM@instpresent\else
    \ClassWarningNoLine{\@classname}{No institution present for an affiliation}%
    \fi
    \if@ACM@citypresent\else
    \ClassWarningNoLine{\@classname}{No city present for an affiliation}%
    \fi
    \if@ACM@countrypresent\else
        \ClassWarningNoLine{\@classname}{No country present for an affiliation}%
    \fi
}
\makeatother

\usepackage{subfigure}
\usepackage{bm}
\usepackage{multirow}
\usepackage{balance}
\usepackage{algorithmicx,algorithm}
\usepackage{algpseudocode}
\usepackage{amsmath}
\usepackage{array}
\usepackage{dcolumn}
\usepackage{enumitem}
\usepackage{soul}
\usepackage{color}

\AtBeginDocument{%
	\providecommand\BibTeX{{%
			\normalfont B\kern-0.5em{\scshape i\kern-0.25em b}\kern-0.8em\TeX}}}

\copyrightyear{2023}
\acmYear{2023}
\setcopyright{acmlicensed}\acmConference[KDD '23]{Proceedings of the 29th ACM SIGKDD Conference on Knowledge Discovery and Data Mining}{August 6--10, 2023}{Long Beach, CA, USA}
\acmBooktitle{Proceedings of the 29th ACM SIGKDD Conference on Knowledge Discovery and Data Mining (KDD '23), August 6--10, 2023, Long Beach, CA, USA}
\acmPrice{15.00}
\acmDOI{10.1145/3580305.3599811}
\acmISBN{979-8-4007-0103-0/23/08}

\begin{document}

	\title[Applied Data Science Track Paper]{DRL4Route: A Deep Reinforcement Learning Framework for Pick-up and Delivery Route Prediction}
 
	\author{Xiaowei Mao$^{\dag}$}
    \author{Haomin Wen$^{\dag}$}
    \affiliation{\institution{School of Computer and Information Technology,\\ Beijing Jiaotong University, Beijing, China}}
    \affiliation{\institution{Cainiao Network, Hangzhou, China}}
    % \email{maoxiaowei@bjtu.edu.cn}
    % \email{wenhaomin@bjtu.edu.cn}
    
    \author{Hengrui Zhang}
    \author {Huaiyu Wan}
    \authornote{Corresponding author;$^{\dag}$Equal contribution}
    \affiliation{\institution{School of Computer and Information Technology,\\ Beijing Jiaotong University, Beijing, China}}
    \affiliation{\institution{Beijing Key Laboratory of Traffic Data Analysis and Mining, Beijing, China}}
    % \email{18112037@bjtu.edu.cn}
    % \email{hywan@bjtu.edu.cn}
    
    \author{Lixia Wu}
    \author{Jianbin Zheng}
    \author{Haoyuan Hu}
    \affiliation{\institution{Cainiao Network, Hangzhou, China}}

    \author{Youfang Lin}
    \affiliation{\institution{School of Computer and Information Technology,\\ Beijing Jiaotong University, Beijing, China}}
    \affiliation{\institution{Beijing Key Laboratory of Traffic Data Analysis and Mining, Beijing, China}}

 \renewcommand{\shortauthors}{Xiaowei Mao et al.}

\begin{abstract}
    \par Pick-up and Delivery Route Prediction (PDRP), which aims to estimate the future service route of a worker given his current task pool, has received rising attention in recent years. Deep neural networks based on supervised learning have emerged as the dominant model for the task because of their powerful ability to capture workers' behavior patterns from massive historical data. Though promising, they fail to introduce the non-differentiable test criteria into the training process, leading to a mismatch in training and test criteria. Which considerably trims down their performance when applied in practical systems. To tackle the above issue, we present the first attempt to generalize Reinforcement Learning (RL) to the route prediction task, leading to a novel RL-based framework called  DRL4Route. It combines the behavior-learning abilities of previous deep learning models with the non-differentiable objective optimization ability of reinforcement learning. DRL4Route can serve as a plug-and-play component to boost the existing deep learning models. Based on the framework, we further implement a model named  DRL4Route-GAE for PDRP in logistic service. It follows the actor-critic architecture which is equipped with a Generalized Advantage Estimator that can balance the bias and variance of the policy gradient estimates, thus achieving a more optimal policy. Extensive offline experiments and the online deployment show that  DRL4Route-GAE improves Location Square Deviation (LSD) by 0.9\%-2.7\%, and Accuracy@3 (ACC@3) by 2.4\%-3.2\% over existing methods on the real-world dataset.

	\end{abstract}
	
	


	\settopmatter{printfolios=true}
	\maketitle 

\pagenumbering{gobble} %remove page number
\section{Introduction} \label{sec:intro}

\par \par Pick-up and delivery services, such as logistics and food delivery, are undergoing explosive development by greatly facilitating people's life. A crucial task in those services is  Pick-up and Delivery Route Prediction (PDRP), which aims to estimate the future service route of a worker given his unfinished tasks. PDRP has received rising attention in recent years from both academia and industry, for it serves as a preliminary task for many downstream tasks in the service platforms (e.g., Cainiao\footnote{https://global.cainiao.com/}, Jingdong\footnote{https://www.jdl.com/} and GrabFood\footnote{https://www.grab.com/sg/food/}). For example, arrival time estimation takes the route prediction result as input, since the arrival time of a task is highly relevant to its order in the worker's entire service route. For another example, the route prediction results can be fed into the dispatching system, to improve efficiency by assigning new orders located or nearby the route of a worker. In light of the above examples, accurate route predictions can certainly improve the user's experience, as well as save the cost for the service providers. 

 \begin{figure}[t]%hbtp
		\centering
		\includegraphics[width=1 \columnwidth]{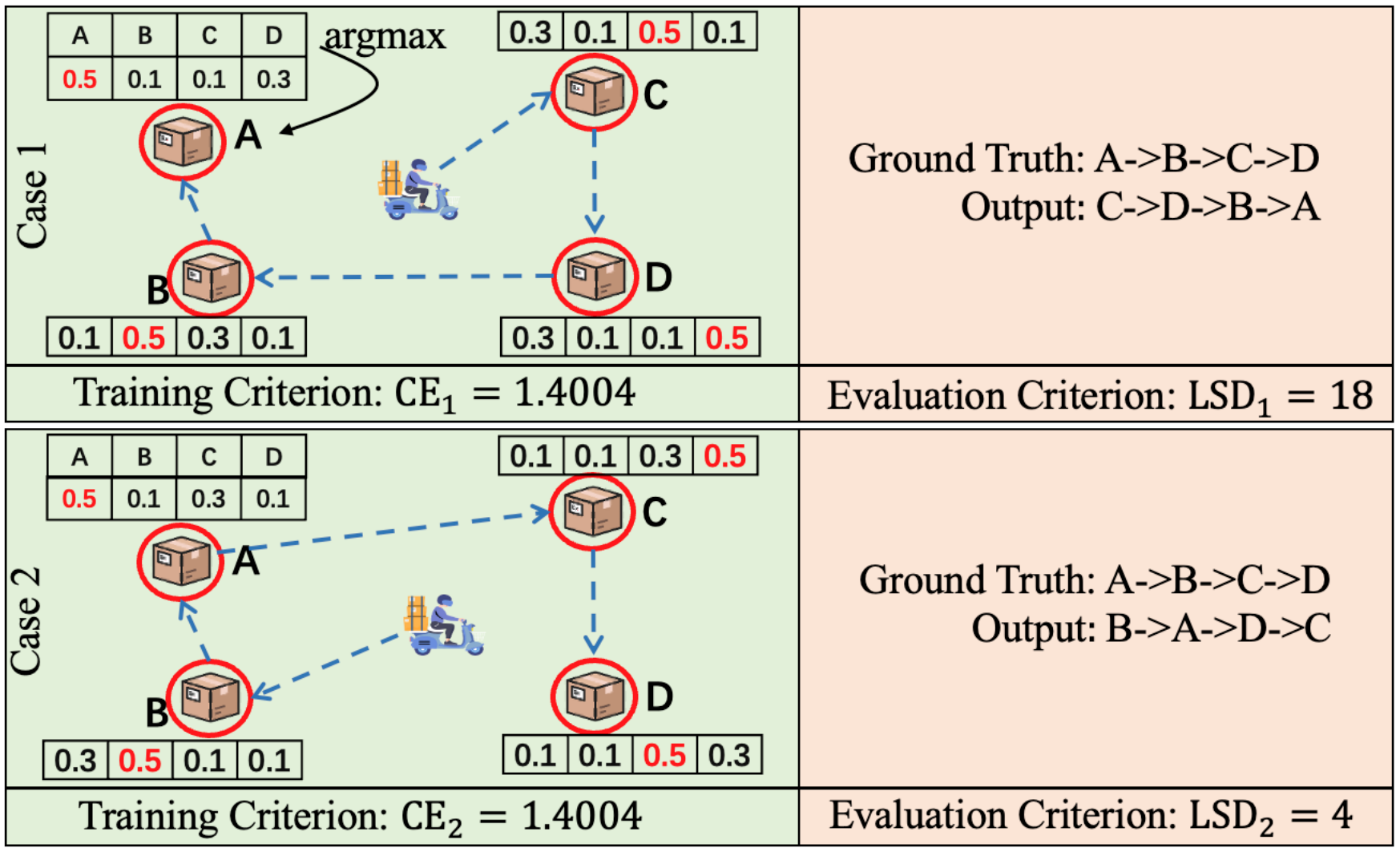}
		\captionsetup{font={small}}
            \vspace{-1em}
		\caption{Illustration of mismatch between the training and test objectives. The vector near each location is the transition probability corresponding to A, B, C, and D.}
      
		\label{fig:introduction}
        \vspace{-1.5 em}
\end{figure}

\par Several learning-based methods \cite{e_le_me, wen2021package, gao2021deep, wen2022graph2route} have been proposed for PDRP tasks. Those methods typically resort to deep neural networks to learn workers' routing patterns in a supervised manner from their massive historical behaviors. OSquare \cite{e_le_me} solves the route prediction problem in food delivery. It converts the route prediction into a next-location prediction problem and outputs the whole route step by step with an XGBoost \cite{chen2016xgboost}. FDNET \cite{gao2021deep} proposes a route prediction module based on the RNN and the attention mechanism to predict the worker's route. In the logistics field, DeepRoute \cite{wen2021package} utilizes a Transformer-like encoder and an attention-based recurrent decoder to predict workers' future pick-up routes. Graph2Route \cite{wen2022graph2route} proposes a dynamic spatial-temporal graph-based model to precisely capture workers' future routing behaviors, thus facilitating accurate route prediction.

\par Though promising, all the above-mentioned deep learning models suffer from the following limitation: \emph{the training criteria is not the same as the test one}, which considerately trims down their performance when applied in the real-world system. To facilitate understanding, we give a simple illustration in Figure~\ref{fig:introduction}. Specifically, those methods train the model using the Cross-Entropy (CE) as the loss function, while evaluating the model using other measurements, such as Location Square Deviation (LSD) \cite{wen2022graph2route} and Kendall Rank Correlation (KRC) \cite{wen2022graph2route}. Thus leading to a mismatch between the training and test objectives. In Figure~\ref{fig:introduction}, although the two cases produce the same value on the training criteria (i.e., CE), their performance on the test criteria (i.e., LSD) is quite different. Such mismatch limits the ability of a ``well-trained'' model to achieve more promising performance in terms of the test criteria. To this end, we ask for a more effective method that can distinguish the above two cases in the training procedure.

\par An intuitive way is transforming the test criteria as the loss function to update the model via its gradient w.r.t the model's parameters. However, it is not applicable since those test criteria (e.g., LSD, KRC) in this task are non-differentiable. They calculate the dissimilarity between the ground-truth route and the predicted route. And the prediction is generated recurrently by the argmax operation shown in Figure~\ref{fig:introduction} (i.e., selecting a candidate task with the maximum output probability at each decoding step), which is unfortunately non-differentiable as elaborated in previous literature \cite{stoyanov2011empirical, corro2018differentiable}. To this end, a solution that could use the test criteria during the training process remains to be explored. Recently, reinforcement learning has shown promising performance in optimizing a task-specific non-differentiable criterion in various tasks, such as text summarization \cite{paulus2017deep, wang2018reinforced}, machine translation \cite{ranzato2015sequence, kreutzer2018reliability}, image captioning \cite{xu2015show, rennie2017self}, and achieves significant improvements compared to previous supervised learning methods.

\par In this paper, we first provide a new perspective to formulate the PDRP task from the RL perspective, and then propose a training framework (DRL4Route) built on policy-based reinforcement learning for PDRP tasks. Our framework optimizes the deep learning models by policy gradient based on reward calculated by non-differentiable test criteria, to solve the mismatch problem between the training and test criterion. On the basis of the framework, we design a model named DRL4Route-GAE for route prediction in logistics service. DRL4Route-GAE mitigates the high variance and avoids unrealistic rewards in the vanilla policy gradient algorithm \cite{keneshloo2019deep} by adopting the actor-critic architecture, which has a lower variance at the cost of significant bias since the critic is not perfect and trained simultaneously with the actor \cite{bahdanau2016actor}. To create a trade-off between the bias and variance of the gradient estimation in actor-critic approaches, the \underline{G}eneralized \underline{A}dvantage \underline{E}stimation \cite{schulman2015high} is adopted to calculate the advantage approximation and update the loss. In summary, the main contributions of this paper are as follows: 

% \begin{itemize}%[leftmargin=*]
\begin{itemize}[leftmargin=*]
    \item We hit the problem of the PDRP task from a reinforcement learning perspective with the first shot, leading to a novel RL-based framework called DRL4Route. Compared with the paradigms of supervised learning in previous works, DRL4Route can combine the power of RL methods in non-differentiable objective optimization with the abilities of deep learning models in behavior learning.

    \item We further propose a model named DRL4Route-GAE for route prediction in logistics. It utilizes an actor-critic architecture that guides the training by observing the reward at each decoding step to alleviate the error accumulation. And it leverages generalized advantage estimation to calculate the advantage approximation and updates the loss function to create a trade-off between the bias and variance of the gradient estimation.
    \item Extensive offline experiments conducted on the real-world dataset and online deployment demonstrate that our method significantly outperforms other solutions, which improves  Location Square Deviation (LSD) by 0.9\%-2.7\%, and Accuracy@3 (ACC@3) by 2.4\%-3.2\%  over the most competitive baseline.
\end{itemize}

\section{Preliminaries} \label{preliminaries}

\par We first introduce the background and then give the problem definition of the pick-up and delivery route prediction in this section.

\par  The pick-up and delivery service mainly involves three roles, including customers, workers, and the platform. Take the food delivery service as an example, a customer first places an online task (i.e., a food order) on the platform with some requirements (e.g., delivery time and location). Then, the platform will dispatch the task to an appropriate worker. At last, the worker will finish his task one by one. We focus on predicting the service route of a worker given his unfinished tasks. The related concepts are formulated as follows:
\par \textbf{Unfinished tasks}. We define the unfinished task set of worker $w$ at the query time as:
\begin{equation}
    O^w = \{o_i|i = 1,...,n\},
\end{equation}
where $n$ is the number of unfinished tasks at the query time, $o_i$ is the $i$-th task, which is associated with a feature vector $x_i \in {\mathbb R}^{d_{feature}}$, including the distance between the worker's current location and the task's location, and the remaining promised pick-up or estimated delivery time of the task.
\par \textbf{Service route.} A service route can be defined as a permutation of the unfinished task set $O^w$:
\begin{equation}
    Y^w = (y_1, ..., y_{ n }),
\end{equation}
where $y_j \in \{1,...,{n}\}$ and if $j \neq j^{'}$ then $y_j \neq y_{j^{'}}$. For instance, $y_j = i$ means the $j$-th task in the service route is the $i$-th task in the unfinished task set.

\par \textbf{Route Constraints.} Various routing constraints exist in real-world services, such as pick-up-before-delivery constraints \cite{gao2021deep} (i.e., the delivery location of a task can only be accessed after the pick-up location has been visited) and capacity constraints \cite{toth2002vehicle} (i.e., the total weight of items carried by a worker cannot exceed its load capacity. The routing constraints can be represented by a rule set $\mathcal C$, each item corresponds to a specific routing constraint.

\par  \textbf{Problem Definition}. Given the unfinished task set $O^w$ of worker $w$ at the query time, we aim to predict the worker's decision on the orders of the unfinished tasks, which can satisfy the given route constraints $\mathcal C$, formulated as:
\begin{equation}
     {{\mathcal F}_{\mathcal C}}({\mathcal O}^w) = ({\hat{y}_1},{\hat{y}_2} \cdots {\hat{y}_{ m}}). \\
\end{equation}
Note that the number of tasks in the label can be less than that in prediction because of the uncertainty brought by new coming tasks as described in \cite{wen2022graph2route}. Formally, we have the prediction $\hat{Y}^w = (\hat{y}_1, \cdots, \hat{y}_n)$ and the label $Y^w=(y_1, \cdots, y_m)$, and the set of the label is included in the set of the prediction. Let $R_{Y}(i)$ and $R_{\hat{Y}}(i)$ be the order of the $i$-th unfinished task in the label and prediction, respectively. 

\section{Proposed DRL4Route Framework} \label{sec:framework}

\par As we have elaborated in Section~\ref{sec:intro}, traditional supervised models fail to introduce the test criteria into the training process due to the non-differentiable characteristics of test criteria. Thus creating the mismatch problem that considerably harms the model's performance. Targeting this challenge, we revisit the PDRP task and generalize it from the reinforcement learning perspective, leading to our proposed RL-based framework called DRL4Route. It solves the mismatch problem by casting traditional deep learning models as the agent and introducing the test criteria as the reward in reinforcement learning. In this section, we first formulate the RDRP task from the reinforcement learning perspective, then elaborate on the proposed framework.

\subsection{Formulation from the RL Perspective}

\par Technically speaking, predicting the future route can be viewed as a sequential decision-making process, where each task on the route is outputted step by step based on previous decisions. From the reinforcement learning perspective, the sequential decision-making process can be represented by a discrete finite-horizon discounted Markov Decision Process (MDP) \cite{sutton1998introduction}, in which a route prediction agent interacts with the environment over discrete time steps $T$. MDP is formulated as $M = (\emph{S}, \emph{A}, \emph{P}, \emph{R}, s_0, \gamma, T)$, where $\emph{S}$ is the set of states, $\emph{A}$ is the set of actions, $\emph{P}: \emph{S} \times \emph{A} \times \emph{S} \rightarrow {\mathbb R}_{+}$ is the transition probability, $\emph{R}: \emph{S} \times \emph{A} \rightarrow {\mathbb R}$ is the reward function, $s_0: \emph{S} \rightarrow {\mathbb R}_{+}$ is the initial state distribution, $\gamma \in \left[0, 1\right]$ is a discount factor, and $T$ is the total time steps determined by the number of unfinished tasks.
\par Given a state $s_t$ at time $t$, the route prediction agent generates an action (i.e., selecting the next task) by the current policy $\pi_\theta$ parameterized by $\theta$, and receives the task-specific reward $r_t$ based on the test criteria such as LSD. The training goal is to learn the best parameter $\theta^{*}$ of the route prediction agent that can maximize the expected cumulative reward, formulated as:
\begin{equation}
    \theta^{*} = {\arg\max}_{\theta}{\mathbb{E}_{\pi_{\theta}}\left[\mathop\sum\limits_{t=1}^{T}{\gamma^t}r_t\right]},
\end{equation}
where the discount factor $\gamma$ controls the tradeoffs between the importance of immediate and future rewards. We introduce the details of the agent, state, action, reward and state transition probability in the following part. 

\par  \noindent \textbf{Route Prediction Agent.} The route prediction agent is designed to model $\mathcal{F}_{\mathcal C}$. It selects a task from unfinished tasks step by step, which follows an encoder-decoder architecture as in previous works. The encoder of the agent aims to compute comprehensive representations for unfinished tasks, formulated as:
\begin{equation}
    {\mathbf E} = { \textbf{Encoder}} (\mathcal{O}^{w}).
    \label{eq:framework_agent_encoder}
\end{equation}
The decoder computes the predicted route $\hat{Y}$ recurrently based on the embedding matrix ${\mathbf E} = (e_1, e_2, \cdots, e_n)$, where $e_i$ denotes the embedding of the $i$-th unfinished task. It contains several decoder cells that learn the output probability distribution of all unfinished tasks. Abstractly, we have
\begin{equation}
\begin{array}{cc}
     o_t, h_{t+1} = \textbf{DecoderCell}(\mathbf E, {\mathcal C}, h_{t}, \hat{Y}_{1: t-1}),\\
\end{array}
\label{agent_decoder}
\end{equation}
where $h_t$ is the decoder hidden state, and $o_t$ is the output probability distribution at the $t$-th decoding step and $\mathcal C$ represents the route constraints. It is designed to meet the service-dependent mask requirements by masking unfeasible tasks at each decoding step. In practice, one can easily change the service-dependent mask mechanism to accommodate route prediction tasks in different scenarios. In addition, we use $\pi_{\theta_a}$ to denote the policy of the agent parameterized by $\theta_a$.

\par  \noindent \textbf{State.} The state $s_t \in \mathcal{S}$ denotes the environment state at the $t$-th decoding step of an agent. It contains the available information for the agent to take actions at each decoding step, defined as $s_t = (\mathbf{E}, \mathcal{C}, h_t, \hat{Y}_{1:t-1})$, where $\hat{Y}_{1:t-1}$ is the outputted route until the $t$-th decoding step.

\par \noindent \textbf{Action.} An action $a_t \in \mathcal{A}_t$ is defined as selecting a task $\hat{y}_t$ given current task candidates and states. And a joint action $(a_1, \cdots, a_n) \in \mathcal{A}=\mathcal{A}_1 \times \cdots \times \mathcal{A}_n$ instructs a predicted route. Where the action space $\mathcal{A}_t$ specifies the task candidates that the agent can select at the $t$-th step. The action space varies along with the decoding process under the route constraints.

\begin{figure}[t]%hbtp
		\centering
		\includegraphics[width=1 \columnwidth]{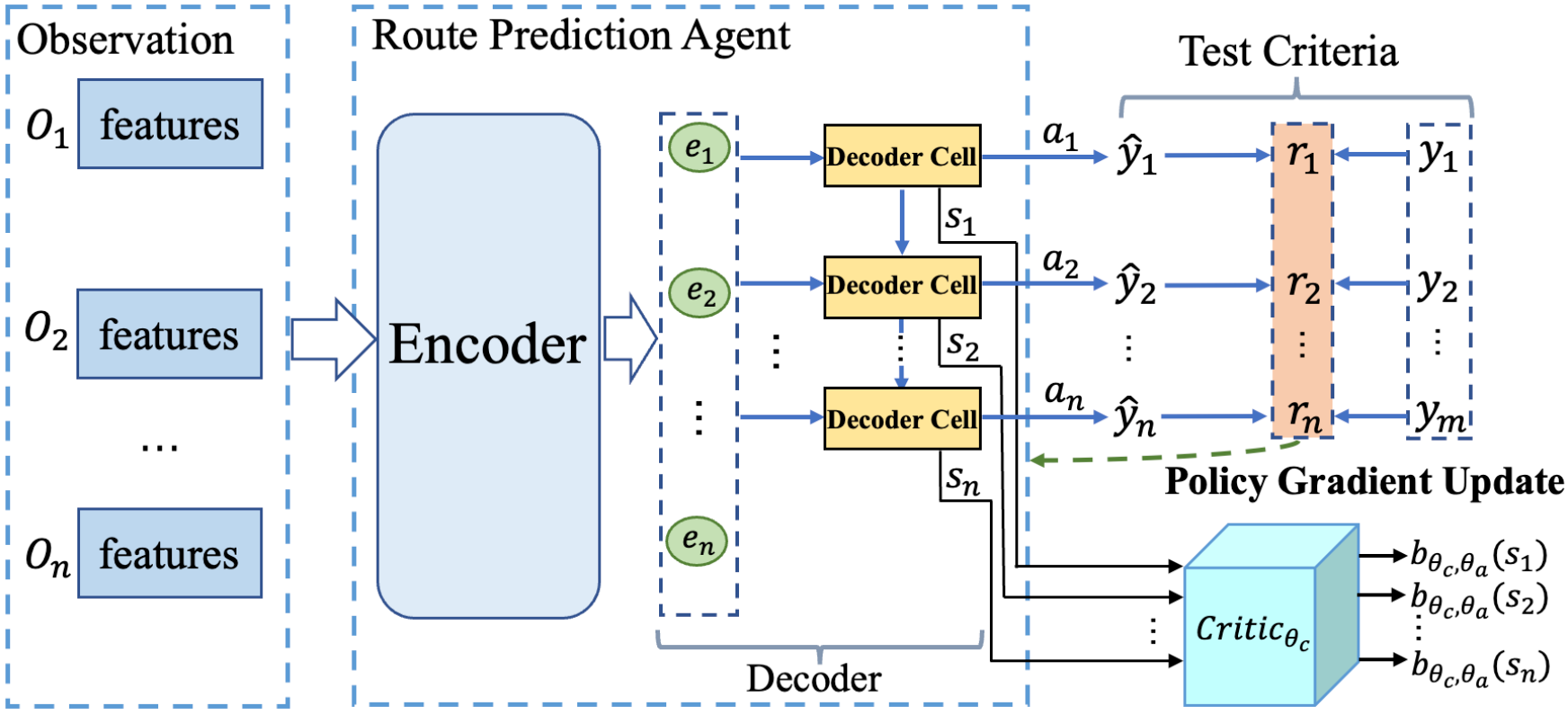}
		\captionsetup{font={small}}
		\caption{{DRL4Route Framework.}}
            \vspace{-2em}
		\label{fig:framework}	
\end{figure}

\par \noindent \textbf{Reward.} The reward is defined based on the test criteria to align the training and test objectives. To ease the presentation, here we choose Location Square Deviation (LSD) in Eqution~\ref{eq:lsd_rl} as an example. 
\begin{equation}
    \begin{aligned}
        {\rm LSD}(p, q)&=(p - q)^2, \\
    \end{aligned}
    \label{eq:lsd_rl}
\end{equation}
where $p, q \geq 0$ represents the order of the task in the route. LSD measures the deviation between the prediction and label, and a smaller LSD indicates better performance. 
\par There are four cases for the reward at each decoding step: case 1) If the outputted task is not in the label and outputted within the first $m$ tasks, where $m$ is the number of tasks in the label, it creates a deviation between the label and prediction; case 2) If the outputted task is not in the label and outputted after the first $m$ tasks, it will not be considered in evaluation since it is influenced by new coming tasks as described in \cite{wen2021package}; case 3) If the outputted task is in the label and not outputted in the right order, we calculate the deviation directly; case 4) If the outputted task is in the label and outputted in the right order, we obtain a reward $\overline{R}$, which is a hyper-parameter to control the scale of the cumulative reward. Thus the reward $r_t$ at the $t$-th decoding step is formulated as:
\begin{equation}
r_t = \left\{ 
{\begin{array}{*{20}{l}}
    -\mathrm{LSD}{(m + 1, t)}&  \hat{y}_t \notin Y, t \leq m, ({\rm case~1})\\
	0 &  \hat{y}_t \notin Y, t > m, ({\rm case~2})\\
	  - \mathrm{LSD}{(R_{\hat{Y}}({\hat{y}_t}) + 1, t)}&  \hat{y}_t \in Y, t \neq \hat{y}_t, ({\rm case~3}) \\
    \overline{R}, & \hat{y}_t \in Y, t = \hat{y}_t, ({\rm case~4})
	\end{array}} \right.
	\label{eq:attention}
\end{equation}

\par \noindent \textbf{State transition probability.} $P(s_{t+1}|s_t, a_t): \emph{S} \times \emph{A} \times \emph{S} \rightarrow {\mathbb R}_{+}$ is the transition probability from state $s_t$ to $s_{t+1}$ if action $a_t$ is taken at $s_t$. In the framework, the environment is deterministic, indicating that the state $s_{t+1}$ transited from state $s_t$ after taking action $a_t$ is determined.

\par The DRL4Route framework is shown in Figure~\ref{fig:framework}. Our framework aims to update the parameters of the agent by policy gradient based on the reward defined by the test criteria. Thus we can optimize the model by non-differentiable objective and achieve more accurate route prediction. Intuitively, if the agent takes action to select a task that meets the test criteria, the reward would be high. Thus, the action is encouraged, and the parameters of the agent are updated to the direction of making such actions. If a bad action is made, the reward would be low, and the agent is updated to the direction of avoiding making such actions.

\subsection{Reinforcement-Guided Route Prediction}

\par \noindent \textbf{Policy-based Reinforcement Learning.}  In the above section, we have formulated the route prediction task as a Markov Decision Process. At each decoding step $t$, the route prediction agent observes the environment state $s_t = (\mathbf{E}, \mathcal{C}, h_t, \hat{Y}_{1:t-1})$, then samples an action from the policy $\pi_{\theta_a}$. After that, the output task at the $t$-th decoding step is determined and the agent receives a reward $r_t$. The goal of the training is optimizing the parameters of the agent to maximize the expected cumulative reward, which is converted to a loss defined as the negative expected cumulative reward of the full sequence:  
\begin{equation}
{\mathcal L_{\theta_a}}=- \mathbb{E}_{\pi_{\theta_a}}[r(\hat{y}_1,\cdots,\hat{y}_n)],
\label{eq_reinforce_loss}
\end{equation}
where $\hat{y}_1,\cdots,\hat{y}_n \sim {\pi_{\theta_a}}$ and they are sampled from the policy ${\pi_{\theta_a}}$. $r(\hat{y}_1,\cdots,\hat{y}_n)$ is the reward associated with the predicted route $(\hat{y}_1,\cdots,\hat{y}_n)$. In practice, we can approximate the expectation in Equation \ref{eq_reinforce_loss} with $N$ sampled routes from the agent's policy $\pi_{\theta_a}$. The derivative for loss ${\mathcal L}_{\theta_a} $ is formulated as:

\begin{equation}
{\nabla_{\theta_a}\mathcal L_{\theta_a}}=- {\frac{1}{N}} \mathop\sum\limits_{i=1}^{N} [\nabla_{\theta_a} \mathrm{log} \pi_{\theta_a} (\hat{y}_{i,1},\cdots,\hat{y}_{i,n})r(\hat{y}_{i,1},\cdots,\hat{y}_{i,n})].
\label{eq_derivative_reinforcement}
\end{equation}
This algorithm is called REINFORCE \cite{williams1992simple} and is a basic policy-based algorithm for route prediction. Technically, REINFORCE can only observe the reward after the full sequence of actions is sampled \cite{8801910}. In that case, the model is forced to wait until the end of the sequence to observe its performance, thus resulting in error accumulation and high variability of the policy gradient estimates \cite{8801910}. 

\par \noindent \textbf{Actor-Critic based Reinforcement Learning.} To tackle the above-mentioned problem, we utilize an Actor-Critic architecture \cite{sutton1998introduction} for the training process. It reduces the variance of the policy gradient estimates, by providing reward feedback of a given action at each time step. Specifically, the ``Critic'' estimates the state-value function (i.e., $V$ function) that evaluates how good it is for a certain state. It also learns a state-action value function (i.e., $Q$ value) that evaluates the benefits of taking a certain action in a certain state. The ``Actor'' is essentially the route prediction agent we want to learn, which updates the policy distribution in the direction suggested by the Critic. Under the policy $\pi_{\theta_a}$, we define the values of the state-action pair $Q(s_t, a_t)$ and the value $V(s_t)$ as follows:

\begin{equation}
    Q_{\pi_{\theta_a}}(s_t, a_t) = \mathbb{E}_{\pi_{\theta_a}}\left[r(\hat{y}_t,\cdots,\hat{y}_n)|s=s_t,a=a_t\right],
\end{equation}
\begin{equation}
    V_{\pi_{\theta_a}}(s_t) = \mathbb{E}_{a_t\sim {\pi_{\theta_a}}(s_t)}\left[Q_{\pi_{\theta_a}}(s_t, a = a_t)\right].
\end{equation}
The state-action value function ($Q$-function) can be computed recursively with dynamic programming:
\begin{equation}
    Q_{\pi_{\theta_a}}(s_t, a_t) = \mathbb{E}_{s_{t+1}}\left[r_t+ \gamma\mathbb{E}_{a_{t+1} \sim {\pi_{\theta_a}}(s_{t+1})}\left[Q_{\pi_{\theta_a}}(s_{t+1}, a_{t+1}) \right] \right].
\end{equation}

A critic network with trainable parameters $\theta_c$ is utilized to approximate the value function $V_{\pi_{\theta_a}}(s)$. In our framework, the critic shares parameters with the actor to fully leverage the spatio-temporal information of unfinished tasks and accurately estimate the state function. The predicted value function is then used to estimate the advantage function $A_{\pi_{{\theta_{a}}}}$:   
\begin{equation}
    \resizebox{1 \linewidth}{!}{
        \begin{math}
        \begin{aligned}
            A_{\pi_{{\theta_{a}}}}(s_t, a_t) &= Q_{\pi_{\theta_{a}}}(s_t, a_t) - V_{\pi_{{\theta_{a}}}}(s_t) \\
              &=  r_t + \gamma\mathbb{E}_{s_{t+1} \sim {\pi_{\theta_{a}}}(s_{t+1}|s_t)}\left[V_{\pi_{{\theta_{a}}}}(s_{t+1}) \right] - V_{\pi_{{\theta_{a}}}}(s_t). \\
        \end{aligned}
         \end{math}
    }
    \label{eq_advantage}
\end{equation}
Inspired by \cite{8801910}, we could sample an unfinished task set to calculate the expectation of the value function in state $s_{t+1}$, and approximate the advantage function to accelerate the calculation as follows:
\begin{equation}
    \resizebox{0.7 \linewidth}{!}{
        \begin{math}
        \begin{aligned}
            A_{\pi_{{\theta_{a}}}}(s_t, a_t) \approx r_t + \gamma V_{\pi_{{\theta_{a}}}}({s_{t+1}}) - V_{\pi_{\theta_{a}}}(s_t). \\
        \end{aligned}
        \end{math}
    }
    \label{eq_sample_advantage}
\end{equation}
Intuitively, the value function $V$ measures how good the policy could be when it is in a specific state $s_t$. The $Q$ function measures the value of choosing a particular action when we are in that state. Based on these two functions, the advantage function $A_{\pi_{\theta_{a}}}$ captures the relative superiority of each action by subtracting the value function $V$ from the $Q$-function.
\par We sample multiple unfinished task sets for the advantage function calculation to reduce the variance of the gradient estimates. In the Actor-Critic architecture, the actor provides samples for the critic network, and the critic returns the value estimation to the actor, and finally, the actor uses these estimations to calculate the advantage approximation function and update the loss according to the following equation:
% \begin{equation}
% {\mathcal L_{{\theta_a}}}={\frac{1}{N}} \mathop\sum\limits_{i=1}^{N}\sum_{t \in  T} {{\rm log}(p}(a_{i, t}|{\theta_a}))A_{{\theta_a}}(s_{i, t}, a_{i, t}).
% \label{eq_loss_reinforce}
% \end{equation}
\begin{equation}
{\mathcal L_{{\theta_a}}}={\frac{1}{N}} \mathop\sum\limits_{i=1}^{N}\sum_{t \in  T}{\rm log}{{\pi}_{\theta_a}(a_{i, t}|s_{i, t})}{A_{\pi_{\theta_{a}}}}(s_{i, t}, a_{i, t}),
\label{eq_loss_reinforce}
\end{equation}
where ${\pi}_{\theta_a}$ is the policy function modeled by the actor network. The critic is a function estimator that tries to estimate $r(\hat{y}_{t},\cdots,\hat{y}_{n})$ for the model at each decoding step $t$, the predicted value $b_{\theta_c, \theta_a}(s_t)$ of the critic is called the ``baseline''. Training the critic is essentially a regression problem, and the value function is trained using a robust loss \cite{girshick2015fast} which is less sensitive to outliers than $L_2$ loss:
\begin{equation}
{\mathcal L_{\theta_c}}={\frac{1}{N}} \mathop\sum\limits_{i=1}^{N}\sum_{t\in T} \mathrm{smooth}{L_1}({b_{\theta_c, \theta_a}}(s_{i, t}) - r(\hat{y}_{i,t},\cdots,\hat{y}_{i,n})),
\label{eq_loss_critic}
\end{equation}
in which $\mathrm{smooth}{L_1}$ is defined as
\begin{equation}
\mathrm{smooth}{L_1}(x) = 
\left\{ 
{\begin{array}{*{20}{l}}
    0.5x^2&  |x|<1,\\
	|x|-0.5& \mathrm{otherwise}.\\
	\end{array}} \right.
\label{l1_loss}
\end{equation}
\par In summary, DRL4Route works as follows: if the route prediction agent outputs a task that matches its actual order in the route, the reward measured by the test criteria would be high, which will encourage the action, and vice versa.

\par \noindent \textbf{Training.} In the beginning, we conduct pre-training to optimize the parameters of the route prediction agent by maximum-likelihood objective, specifically, minimize the following cross-entropy loss:
 \begin{equation}
{\mathcal L_{CE}}=- {\frac{1}{N}}\mathop\sum\limits_{i=1}^{N}\sum_{t\in T} {{\rm log}(P}(y_{i, t}|\theta_a)),
\label{eq_CE_loss}
\end{equation}
 After that, we conduct the joint training for the actor and the critic by the following loss function:
\begin{equation}
{\mathcal L}_{\theta_{ac}} = \alpha_{\theta_c}{\mathcal L_{\theta_c}} + \alpha_{\theta_a}{\mathcal L_{\theta_a}} + \alpha_{CE}{\mathcal L_{CE}},
\label{eq:eq_all}
\end{equation}
where $\theta_{ac} = \theta_a \cup \theta_c$ represents the parameters of both the actor and the critic. $\alpha_{\theta_c}, \alpha_{\theta_a}, \alpha_{CE}$ are hyper-parameters to control the weight of different loss functions.  
\par \noindent \textbf{Prediction.} When used for prediction, the actor takes all the unfinished tasks as input, and predicts the whole service route step by step, the task to select at each decoding step $t$ is given by:
\begin{equation}
\hat{y}_t={\arg\max}{_k}p_k^t,
\label{eq:predict}
\end{equation}
where $p_k^t$ is the output probability of task $k$ at the $t$-th step.

\section{DRL4Route-GAE: Model for PDRP TASK in Logistics Pick-up Service} \label{sec:model}
\par Based on the DRL4Route framework, we further propose a model called DRL4Route-GAE for logistics pick-up service to demonstrate the effectiveness of the framework. DRL4Route-GAE is equipped with a transformer encoder to generate a spatial-temporal representation of the unfinished tasks and an attention-based recurrent decoder to model the decision process of the worker. The training of the model is guided by policy gradient in an actor-critic manner, so we can optimize the model based on non-differentiable test criteria to solve the problem of mismatch between the training and test objectives. In addition, we leverage generalized advantage estimation to calculate the advantage approximation to create a trade-off between the bias and variance of the gradient estimation in the actor-critic approach. Thus we can achieve a more optimal policy and better results. 
\subsection{Actor-Critic Network} \label{sec:Actor}

\subsubsection{Actor network.}
\par The actor network follows the encoder-decoder architecture. The encoder layer adopts spatial-temporal constraints of unfinished tasks to produce their representations. The decoder layer selects task $\hat{y_t}$ from the unfinished task set at the $t$-th decoding step. And the selected task is used as input for the next decoding step. Finally, the whole predicted route is generated recurrently. The probability of an output service route $\hat{Y}^w$ is expressed as a product of conditional probabilities according to the chain rule:
\begin{equation}
    \begin{aligned}
         P({\hat{Y}^w}|\mathcal{O}^w;\theta_a) =\begin{matrix}\prod_{t=1}^{n}\end{matrix} P(\hat{y}_t|\mathcal{O}^w, \mathcal{C}, \hat{Y}_{1:t-1};\theta_a), \\
    \end{aligned}
    \label{eq:output}
\end{equation}
In the logistic pick-up route prediction problem, the route constraints $\mathcal C$ is a simple rule that requires no duplication output (i.e., a task cannot be outputted twice in a single route).
\par \noindent \textbf{Encoder Layer.} Given the feature matrix ${\mathbf{X}} \in \mathbb{R}^{n \times {d_{feature}}}$ of the unfinished tasks, the encoding layer consists of $N$ blocks of transformer is utilized to integrate the spatial-temporal information of unfinished tasks. Each transformer block consists of two sublayers: a multi-head attention (MHA) layer and a feed-forward network (FFN) layer. We denote the embedding produced by block $l \in \{1, \cdots,K\}$ as $e^l_j$. $n_{head}$ is the number of heads. The embedding updating process can be formulated as follows:
\begin{equation}
\begin{array}{cc}
     \hat{e}_j = {\mathrm {BN}}^l(e^{l-1}_j + \mathrm{MHA}^l_j(\mathrm{head}^{l-1}_1,\cdots,\mathrm{head}^{l-1}_{n_{head}})), \\
     e^l_j = {\mathrm {BN}}^l(\hat{e}_j + \mathrm{FFN}(\hat{e}_j)).
\end{array}
\label{eq_embed_update}
\end{equation}
% Then we obtain each task's embedding $e_j$. We also compute an aggregated embedding $\overline{e}={\frac{1}{n}}\mathop\sum\limits_{j=1}^{n}e_j$ to represent the whole unfinished task set. 

\begin{figure}[t]%hbtp
	\centering
		\includegraphics[width=1 \columnwidth]{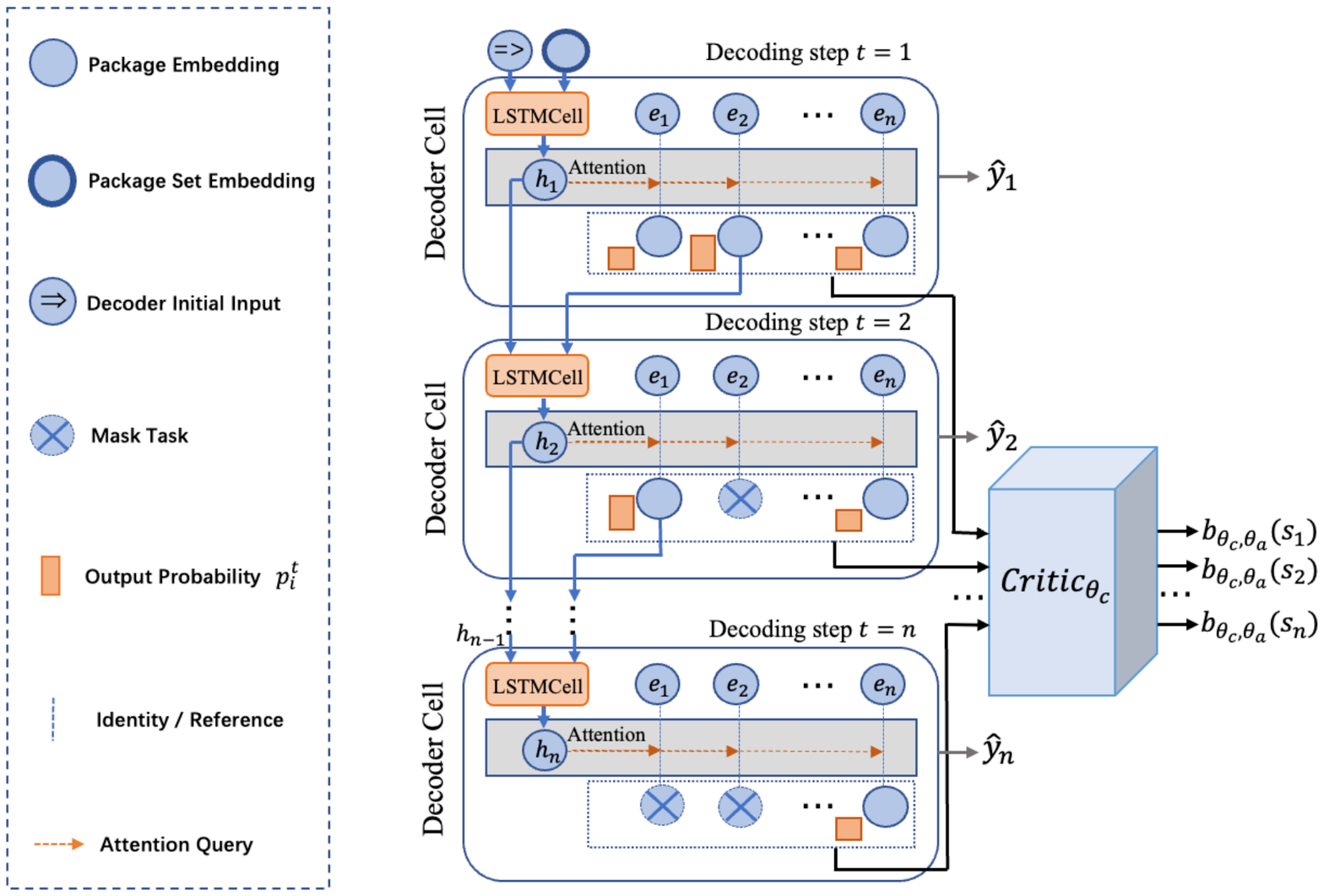}
		\captionsetup{font={small}}
		\caption{Illustration of the decoding process and the critic network. The decoder generates the route recurrently by several decoder cells, with each generates an action based on current state $s_t$. And each state $s_t$ is fed into the critic network to estimate the $V$ function.}
		\label{fig:decoding}
		
\end{figure}
\par \noindent \textbf{Decoder Layer.} An attention-based recurrent decoder layer is leveraged to generate the prediction route. At decoding step $t$, the decoder outputs the task index ${\hat{y}_t}$ based on the embeddings of the encoder layer and the previous outputs of the decoder before step $t$. As shown in Figure~\ref{fig:decoding}, we use LSTM combined with an attention mechanism to compute the output probability $P(\hat{y}_t|\mathcal{O}^w, \mathcal{C}, \hat{Y}_{1:t-1})$ in Equation \ref{eq:output}. The initial input to the decoder consists of all the task embeddings $\{e_1, e_2, \cdots, e_n \}$, the aggregated embedding $\overline{e}={\frac{1}{n}}\mathop\sum\limits_{j=1}^{n}e_j$, and the randomly initialized hidden state of LSTM that denoted as $h_0$.  Then we compute the attention score for all the unfinished tasks at each decoding step $t$, and mask $(u^t_j = -\infty)$ for the tasks which have been output before step $t$ to meet the route constraints:
\begin{equation}
u_j^t = \left\{ 
{\begin{array}{*{20}{l}}
	{{\mathbf v}^T}\tanh ({\mathbf{W}_1}{e_j} + {\mathbf{W}_2}{h_t})&{{\rm{ if }}{\kern 1pt}  j \neq \hat{y}_{t^{'}}, \forall t^{'} \leq t },\\
	{ - \infty }&{{\rm{ otherwise,}}}
	\end{array}} \right.
	\label{eq:attention}
\end{equation}
where ${\mathbf{W}_1}, {\mathbf{W}_2} \in \mathbb{R}^{d_h \times d_h}$ and ${\mathbf v} \in \mathbb{R}^{d_h}$ are learnable parameters. Finally, we get the output probability $p_j^t$ of task $j$ at step $t$ by a softmax layer:
% \begin{equation}
% {p_i^t} = p\left( {{\hat{y}_t} = i~|~\mathcal{O}^w, \mathcal{C}, {{\bf{\hat{Y}}}_{1:t - 1}};\theta_a } \right) = \pi_{\theta_{a}}(u_i^t) =  {\rm softmax}(u_i^t),
% \label{eq:output_prob} 
\begin{equation}
{p_j^t} = P\left( {{\hat{y}_t} = j~|~\mathcal{O}^w, \mathcal{C}, {{\bf{\hat{Y}}}_{1:t - 1}};\theta_a } \right) =  \frac{e^{u^t_j}}{{\begin{matrix}\sum_{k=1}^{n}\end{matrix}}{e^{u^t_k}}}.
\label{eq:output_prob} 
%\nonumber
\end{equation}

\subsubsection{Critic network.}
\par In the policy-based actor-critic architecture, the critic network is expected to give a sound estimate of the state-value function to reduce the variance of the gradient estimation. Thus, the critic network is supposed to grasp comprehensive information about the environment. In our formulation, the environment state includes the spatial-temporal information of unfinished tasks, route constraints, and already outputted tasks in the decoding process. Since the output probability distribution at each decoding step is modeled by the state and contains enough information to evaluate the state-value function, the parameters of the actor are shared with the critic, and the output probability distribution at each decoding step is fed into the critic as shown in Figure~\ref{fig:decoding}. In the implementation, the critic with parameter $\theta_c$ is defined by a feed-forward network as a regression layer and is trained in parallel with the actor network.

\subsection{Generalized Advantage Estimation based Training} \label{sec:train}
On the one hand, actor-critic models usually have low variance due to the batch training and use of critic as the baseline reward. However, actor-critic models are not unbiased. Since they usually leverage neural networks to approximate the $V$ function and use the approximated value to estimate the $Q$ value, the estimated $Q$ value is biased if they fail to accurately approximate the $V$ function. On the other hand, models based on REINFORCE have low bias but can cause high variance since they calculate the expected discounted reward by sampling. To create a trade-off between the bias and variance in estimating the expected gradient of the policy-based loss function, we adopt the generalized advantage estimation \cite{schulman2015high} as follows:
\begin{equation}
    \resizebox{1 \linewidth}{!}{
        \begin{math}
        \begin{aligned}
            A_{\pi_{\theta_a}}^{{\rm GAE}(\gamma, \lambda)}(s_t, a_t) = {\mathop\sum\limits_{t^{'}=t}^{T}}{(\gamma \lambda)}^{(t^{'} - t)}(r(s_{t^{'}}, a_{t^{'}}) + \gamma V_{{\pi}_{\theta_a}}({s_{t^{'}+1}}) - V_{{\pi}_{\theta_a}}(s_{t^{'}})), \\
        \end{aligned}
        \end{math}
    }
    \label{eq:GAE}
\end{equation}
where $\lambda$ controls the trade-off between the bias and variance, such that large values of $\lambda$ yield to larger variance and lower bias, while small values of $\lambda$ do the opposite. There are two special cases of Equation~\ref{eq:GAE}, obtained by setting $\lambda = 0$ and $\lambda = 1$. If $\lambda = 0$, the advantage estimation is given by Equation~\ref{eq_advantage}. In this case, more bias is introduced in the gradient estimation due to assigning more weight to the approximated V function, but the variance is reduced. If $\lambda = 1$, the gradient estimation has a lower bias but has a high variance due to calculating the expected discounted reward by sampling. Overall, the training of DRL4Route-GAE follows that of DRL4Route framework and is illustrated in Algorithm~\ref{algo:Rl2Route}. We first pre-train a route agent (i.e., the actor) by the cross-entropy loss, then jointly optimize it with the critic in the framework.

\floatname{algorithm}{Algorithm}  
\renewcommand{\algorithmicrequire}{\textbf{Input:}}  
\renewcommand{\algorithmicensure}{\textbf{Output:}}  
\begin{algorithm} [t]
	\caption{DRL4Route-GAE.} 
	\begin{algorithmic}[1] %每行显示行号  
		\Require Input sets of unfinished tasks $\mathcal{O}$, and the corresponding ground-truth service route sets $\mathcal{Y}$. The actor network with parameters $\theta_{a}$ and the critic network with parameters $\theta_{c}$.
		\Ensure Trained actor network ${\theta_{a}^*}$.
		\State Initialize $\theta_a$ and $\theta_{c}$;
        \State // Pre-train the actor network using cross-entropy;
        \While{not converged}
        \State Select a batch of size $N$ from $\mathcal{O}$ and $\mathcal{Y}$;
        \State Calculate the loss based on Equation~\ref{eq_CE_loss};
        \State Update the parameters of the actor network $\theta_{a} \gets \theta_{a} + \alpha \nabla_{\theta_{a}}\mathcal{L}_{CE}$;
        \EndWhile
        \State // Train actor-critic model
        \While{not converged}
        \State Select a batch of size $N$ from $\mathcal{O}$, and the corresponding ground-truth service route sets $\mathcal{Y}$;
        \State Sample $N$ sequence of actions based on $\pi_{\theta_a}$;
        \For{$i = 1, ..., N$}
            \For{$t = 1, ..., n$}
            \State Calculate the expected discounted reward:
            \State $r(\hat{y}_{i,t},\cdots,\hat{y}_{i,n}) = \mathop\sum\limits_{t^{'}=t}^{n}r(s_{i, t^{'}}, a_{i, t^{'}})$
            \EndFor 
        \EndFor 
        \State Calculate the loss based on Equation~\ref{eq_loss_reinforce}, \ref{eq_loss_critic}, \ref{l1_loss}, \ref{eq_CE_loss}, \ref{eq:eq_all}, \ref{eq:GAE};
        \State Update parameters of the actor-critic model $\theta_{ac} \gets \theta_{ac} + \alpha \nabla_{\theta_{ac}}\mathcal{L}_{\theta_{ac}}$;
        \EndWhile
		\State return $\theta_{a}^*=\theta_a$;
	\end{algorithmic}
	\label{algo:Rl2Route}
\end{algorithm} 

%\newpage

\section{Experiments}
\par In this section, we present offline as well as online experiments on the real-world dataset to demonstrate the effectiveness of our model.

\subsection{Offline Experiments}
\subsubsection{Datasets.}
\par We evaluate our model based on two real-world logistics pick-up datasets collected from Cainiao\footnote{https://www.cainiao.com/}, one of the largest logistics platforms in China, handling over a hundred million packages per day. The first dataset (Logistics-HZ) contains the pick-up records of 1,117 workers from Jul 10, 2021, to Sept 10, 2021 (90 days) in Hangzhou, China. The second dataset (Logistics-SH) contains the pick-up records of 2,344 workers from Mar 29, 2021, to May. 27, 2021 (60 days) in Shanghai, China. In Logistics-HZ/Logistics-SH, we use the previous 54/36 days as the training set, the last 18/12 days as the test set, and the left days as the validation set. Detailed statistics of the datasets are shown in Table~\ref{tab:data}.

\begin{table}[t]%htbp
	\centering
        % \vspace{-0.5em}
	\caption{Statistics of the Datasets. (Abbr.: ANUT for Average number of unfinished tasks.)}
        \vspace{-0.9em}
	\setlength\tabcolsep{2 pt}
	\resizebox{1 \linewidth}{!}{
		\begin{tabular}{cccccccc}
			\toprule
			Type& Time Range & City & ANUT & \#Workers  & \#Samples \\
			\midrule
			Logistics-HZ &  {07/10/2021 - 10/10/2021}  &Hangzhou & 7 & 1,117 & 373,072\\
			Logistics-SH & {03/29/2021 - 05/27/2021}  & Shanghai & 9 & 2,344  & 208,202\\
			\bottomrule
		\end{tabular}
 		\label{tab:data}
	}
         \vspace{-1.8em}
	\label{datasets}
        
\end{table}

\subsubsection{Baselines.}
\par We implement some basic methods, machine learning methods, and state-of-the-art deep learning models for a comprehensive comparison. 
   
\par \noindent \textbf{Non-learning-based methods}: 
\begin{itemize}
    \item Time-Greedy: A greedy algorithm chooses to take the most urgent  package, regardless of distance requirements and other factors. 
    \item Distance-Greedy: A greedy algorithm chooses to take the nearest  package, regardless of time requirements and other factors.
    \item OR-Tools ~\cite{bello2016neural}: A heuristic algorithm that trades oﬀ optimality for computational cost. We use OR-Tools open source software to find the shortest route for the worker.
\end{itemize}
\par \noindent \textbf{Learning-based models}:
\begin{itemize}
     \item OSquare\cite{e_le_me}: A point-wise ranking algorithm that trains a machine learning model (i.e., LightGBM \cite{DBLP:conf/nips/KeMFWCMYL17}) to output the next location at one step, and the whole route is generated recurrently.
    \item DeepRoute \cite{wen2021package}:  A deep learning algorithm equipped with a Transformer encoder and an attention-based decoder to rank a worker's all unpicked-up packages. 
    \item DeepRoute+ \cite{DeepRoute+}: Based on DeepRoute, it adds an encoding module to model the decision preferences of the workers.
    \item FDNET \cite{gao2021deep}: A deep learning algorithm utilizes the LSTM and the attention mechanism to predict the worker's route in the food delivery system. It introduces the arrival time prediction as the auxiliary task.
    \item Graph2Route \cite{wen2022graph2route}: A deep learning algorithm equipped with a dynamic spatial-temporal graph encoder and a graph-based personalized route decoder. 
\end{itemize}
Moreover, we also implement two methods based on our framework to show its flexibility, including:
\begin{itemize}
     \item DRL4Route-REINFORCE: DRL4Route trained by REINFORCE. The derivative for loss of DRL4Route-REINFORCE is formulated in Equation~\ref{eq_derivative_reinforcement}.
     \item DRL4Route-AC: DRL4Route-GAE trained with $\lambda = 0$ in the generalized advantage estimation.
\end{itemize}
\par For deep learning models, hyperparameters are tuned using the validation set, we train, validate, and test the model at least five times and the average performance is reported. The code is available at https://github.com/maoxiaowei97/DRL4Route.

\subsubsection{Metrics.} {\label{metrics}}
\par Following the setting in \cite{wen2022graph2route}, we introduce a comprehensive indicator system to evaluate the performance from both local and global perspectives. 

\par \noindent  \textbf{KRC}: Kendall Rank Correlation \cite{kendall1938new} is a statistical criterion to measure the ordinal association between two sequences. 
For sequences $\hat{Y}^w$ and $Y^w$, we define the node pair $(i, j)$ to be concordant if and only if both $R_{\hat{Y}}(i) > R_{\hat{Y}}(j)$ and $R_{Y}(i) > R_{Y}(j)$, or both $R_{\hat{Y}}(i) < R_{\hat{Y}}(j)$ and $R_{Y}(i) < R_{Y}(j)$. Otherwise it is said to be discordant.

To calculate this metric, nodes in the prediction are first divided into two sets: i) nodes in label ${{\mathcal V}_{in}} = \{ {\hat y}_i | {\hat y}_i \in {Y^w} \}$, and ii) nodes not in label ${{\mathcal V}_{not}} = \{ {\hat y}_i |  {\hat y}_i \not \in {Y^w} \}$. We know the order of items in $\mathcal{V}_{in}$, but it is hard to tell the order of items in  ${\mathcal V}_{not}$, still we know that the order of all items in $\mathcal{V}_{in}$ are ahead of that in ${\mathcal V}_{not}$. Therefore, we compare the nodes pairs $\{(i,j) | i,j \in {\mathcal V}_{in}~{\rm and}~{i \neq j}\} \cup \{(i,j) | i \in {\mathcal V}_{in} {~\rm and~~} j \in {\mathcal V}_{not} \}$. In this way, the KRC is defined as:

\begin{equation}
	{\rm{KRC}} = \frac{N_c-N_d}{N_c+N_d},
	\label{eq:krc}
\end{equation}
where $N_c$ is the number of concordant pairs, and $N_d$ is the number of discordant pairs.
		
\par \noindent  \textbf{ED:} Edit Distance \cite{nerbonne1999edit} (ED) is an indicator to quantify how dissimilar two sequences $Y^w$ and $\hat{Y}^w$ are to one another, by counting the minimum number of required operations to transform one sequence into another. 

\par \noindent  \textbf{LSD} and \textbf{LMD}: The Location Square Deviation (LSD) and the Location Mean Deviation (LMD) measure the degree that the prediction deviates from the label. The smaller the LSD (or LMD), the closer the prediction to the label. They are formulated as:
\begin{equation}
    \begin{aligned}
        {\rm LSD}&=\frac{1}{m}\sum_{i=1}^{m}(R_{Y}(y_i)-R_{\hat{Y}}({\hat y}_i))^2, \\
    {\rm LMD}&=\frac{1}{m}\sum_{i=1}^{m}|R_{Y}(y_i)-R_{\hat{Y}}({\hat y}_i)|. \\
    \end{aligned}
    \label{eq:lsd_lmd}
\end{equation}

\par \noindent  \textbf{HR@$k$}: Hit-Rate@$k$ is used to quantify the similarity between the top-$k$ items of two sequences. It describes how many of the first $k$ predictions are in the label, which is formulated as follows:
\begin{equation}
    \textbf{\rm HR@}k=\frac{{\hat {Y}}_{[1:k]}\cap {Y}_{[1:k]}}{k}.
    \label{eq_hit_rate}
\end{equation}

\begin{table*}[t]
\centering
\small
\caption{\centering {Experiment Results. Higher KRC, HR@$1$, ACC@$3$ and lower LMD and LSD, ED indicates better performance. }}
%\vspace{-8pt}
\renewcommand\arraystretch{1.1}
\setlength\tabcolsep{2 pt}
\resizebox{1.0 \textwidth}{!}{
\begin{tabular}{c|cccccc|cccccc|cccccc|cccccc}
\hline %\hline
\multirow{3}{*}{Method}
& \multicolumn{12}{|c|}{Logistics-HZ}& \multicolumn{12}{|c}{Logistics-SH} \\
\cline{2-25}
& \multicolumn{6}{|c|}{$n\in (0,11]$}& \multicolumn{6}{|c|}{$n\in (0,25]$}& \multicolumn{6}{|c|}{$n\in (0,11]$}& \multicolumn{6}{|c}{$n\in (0,25]$}\\
\cline{2-25}
  & HR@1 & ACC@3 & KRC & LMD & LSD & ED & HR@1 & ACC@3 & KRC & LMD & LSD & ED & HR@1 & ACC@3 & KRC & LMD & LSD & ED & HR@1 & ACC@3 & KRC & LMD & LSD & ED\\
\hline \hline
Time-Greedy & 33.15 & 20.32 & 41.92 & 1.70 & 6.85 & 1.78 & 32.26 & 19.52 & 40.80 & 1.80 & 7.81 & 2.20 & 26.37 & 13.62 & 37.76 & 2.30 & 11.54 & 2.41 & 25.19 & 12.47 & 35.44 & 2.45 & 12.84 & 3.19 \\ \hline
Distance-Greedy & 51.68 & 34.13 & 52.68 & 1.27 & 5.02 & 1.48 & 51.27 & 33.13 & 51.82 & 1.36 & 5.73 & 1.84 & 45.98 & 26.09 & 51.29 & 1.72 & 8.45 & 2.01 & 45.43 & 24.52 & 49.72 & 1.84 & 9.27 & 2.66 \\ \hline
OR-Tools & 52.72 & 35.14 & 53.93 & 1.23 & 4.68 & 1.46 & 52.20 & 34.01 & 52.98 & 1.31 & 5.42 & 1.83 & 48.59 & 28.04 & 54.30 & 1.54 & 6.87 & 1.95 & 47.81 & 26.26 & 52.60 & 1.67 & 7.73 & 2.61 \\ \hline
OSquare & 54.00 & 33.10 & 58.50 & 1.16 & 3.74 & 1.50 & 53.50 & 32.00 & 57.70 & 1.22 & 4.36 & 1.89 & 47.03 & 24.24 & 55.20 & 1.52 & 6.01 & 2.05 & 46.32 & 22.55 & 53.58 & 1.64 & 6.88 & 2.74 \\ \hline
FDNET & 52.76 & 33.22 & 55.47 & 1.18 & 4.14 & 1.46 & 52.18 & 32.31 & 54.66 & 1.26 & 4.72 & 1.84 & 49.50 & 27.73 & 55.75 & 1.60 & 7.59 & 1.96 & 48.81 & 25.91 & 54.08 & 1.72 & 8.38 & 2.62 \\ \hline
DeepRoute & 54.76 & 34.64 & 58.61 & 1.10 & 3.71 & 1.45 & 54.21 & 33.63 & 57.79 & 1.17 & 4.28 & 1.83 & 51.87 & 28.35 & 59.07 & 1.42 & 5.98 & 1.96 & 50.88 & 26.46 & 57.31 & 1.55 & 6.81 & 2.62 \\ \hline
DeepRoute+ & 55.42 & 35.63 & 59.32 & 1.08 & 3.65 & 1.44 & 54.91 & 34.58 & 58.57 & 1.16 & 4.20 & 1.82 & 52.03 & 28.75 & 59.80 & 1.39 & 5.73 & 1.94 & 51.14 & 26.87 & 58.09 & 1.52 & 6.54 & 2.60 \\ \hline
Graph2Route & \underline{56.45} & 36.12 & 60.63 & \underline{1.05} & \underline{3.47} & \underline{1.43} & \underline{55.99} & 34.94 & 59.87 & \underline{1.13} & 4.04 & 1.81 & \underline{52.53} & 29.25 & \underline{61.22} & \underline{1.34} & \underline{5.21} & \underline{1.92} & 51.56 & 27.28 & 59.45 & \underline{1.46} & \underline{6.02} & \underline{2.58} \\ \hline \hline
DRL4Route-REINFORCE & 55.88 & 35.74 & 60.57 & 1.05 & 3.47 & 1.44 & 55.39 & 34.61 & 59.85 & 1.13 & \underline{4.03} & 1.81 & 52.30 & 29.02 & 59.97 & 1.38 & 5.40 & 1.93 & 51.41 & 26.94 & 58.25 & 1.50 & 6.23 & 2.59 \\ \hline
DRL4Route-AC & 56.36 & \underline{36.16} & \underline{60.86} & 1.05 & 3.47 & 1.43 & 55.88 & \underline{35.00} & \underline{60.13} & 1.13 & 4.03 & \underline{1.80} & 52.51 & \underline{29.27} & 61.14 & 1.34 & 5.26 & 1.93 & \underline{51.63} & \underline{27.32} & \underline{59.49} & 1.46 & 6.04 & 2.58 \\ \hline
DRL4Route-GAE & \textbf{57.72} & \textbf{37.23} & \textbf{61.47} & \textbf{1.03} & \textbf{3.44} & \textbf{1.41} & \textbf{57.23} & \textbf{36.03} & \textbf{60.76} & \textbf{1.11} & \textbf{3.98} & \textbf{1.78} & \textbf{53.08} & \textbf{29.95} & \textbf{61.80} & \textbf{1.32} & \textbf{5.08} & \textbf{1.91} & \textbf{52.24} & \textbf{28.02} & \textbf{60.13} & \textbf{1.44} & \textbf{5.86} & \textbf{2.56} \\ \hline
Improvement & 2.2\% & 3.1\% & 1.4\% & 1.9\% & 0.9\% & 1.4\% & 2.2\% & 3.2\% & 1.5\% & 1.8\% & 1.5\% & 1.7\% & 1.0\% & 2.4\% & 0.9\% & 1.5\% & 2.5\% & 0.5\% & 1.3\% & 2.7\% & 1.1\% & 1.4\% & 2.7\% & 0.8\% \\ \hline
 %\hline
\end{tabular}
}

\label{tab:Table1}
\end{table*}

\par \noindent  \textbf{ACC@$k$}: ACC@$k$ is a stricter measure than HR@$k$ for calculating the local similarity of two sequences.
It is used to indicate whether the first $k$ predictions ${\hat {Y}}_{[1:k]}$ are consistent with those in the label ${Y}_{[1:k]}$.
It is formulated as:
\begin{equation}
    \textbf{\rm ACC@}k= \prod_{i=0}^k{\mathbb{I}}({\hat y}_i, y_i), 
\end{equation}
where ${\mathbb I}(\cdot)$ is the indicator function, and $\mathbb I({\hat y}_i, y_i)$ equals 1 if ${\hat y}_i = y_i$ else 0.

\par In summary, KRC, ED, LSD, and LMD measure the overall similarity of the predicted route and the label route, while HR@$k$ and ACC@$k$ calculate thei similarity from the local perspective. Higher KRC, HR@$k$, ACC@$k$, and lower ED, LSD, LMD mean better performance of the algorithm. 

\subsubsection{Results.} {\label{results}}

We evaluate different algorithms under the Logistics-SH and Logistics-HZ datasets. 
Table \ref{tab:Table1} shows the performance of our  DRL4Route-GAE model achieves significant performance improvements comparing with all the baselines.
On Logistics-HZ (length 0-25), it improves the performance of Graph2Route by 3.2\% at ACC@3,  by 1.8\% at LMD, and by 2.2\% at HR@1.
On Logistics-SH (length 0-25), it improves the performance of Graph2Route by 2.7\% at ACC@3,  by 1.4\% at LMD, and by 2.7\% at LSD.

Time-Greedy and Distance-Greedy perform poorly in basic methods because they only consider time or distance information. Or-Tools tries to find the shortest path for the worker, which is an improvement over the previous two algorithms.

The superiority of DeepRoute over FDNET in the deep model lies in the powerful encoding capabilities brought by the transformer encoder. 
Graph2Route performs best among methods that do not adopt the policy gradient techniques because it uses a dynamic spatial-temporal graph encoder and a graph-based personalized route decoder. At the same time, Graph2Route can capture the decision context information.

Better results were obtained using the  DRL4Route-REINFORCE method on top of DeepRoute, especially on the LSD metric. The reason is that the optimization of the  DRL4Route-REINFORCE is directly based on the evaluation metric, thus solving the problem of inconsistent training and testing objectives generated by the cross-entropy based methods. 
DRL4Route-AC achieves better results than DRL4Route-REINFORCE because the DRL4Route-AC method adopts a critic network to estimate the value function, reducing the policy gradient estimation variance. It also alleviates the error accumulation problem by allowing the model parameters to be updated at each decoding step according to the action. 
DRL4Route-GAE achieves better results than DRL4Route-AC by using the Generalized Advantage Estimator to balance the bias and variance of the policy gradient estimates in the AC method.

\begin{figure}[hbtp]%hbtp
		\centering
		\includegraphics[width=0.7 \columnwidth]{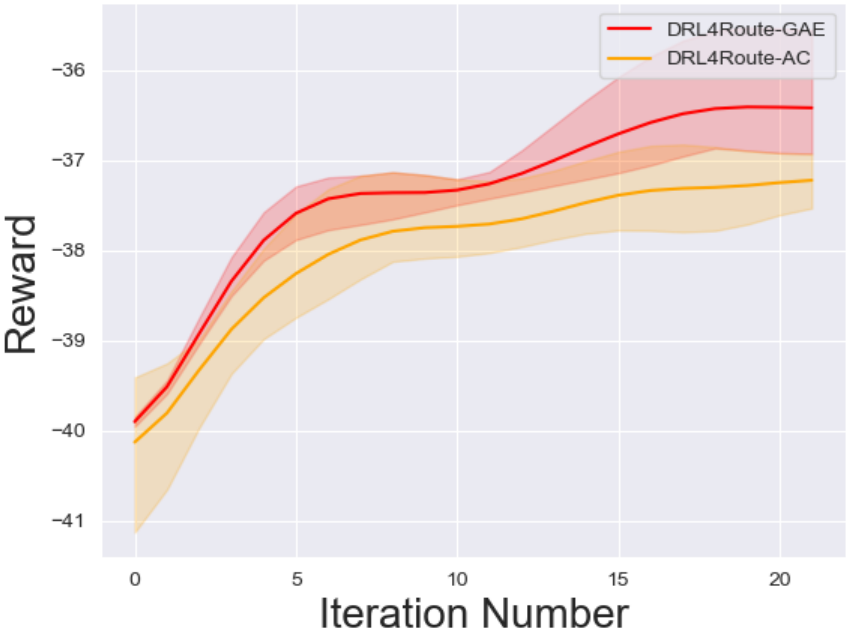}
		\captionsetup{font={small}}
        \vspace{-0.5em}
		\caption{Reward curves for the training process of the DRL4Route-AC and DRL4Route-GAE..}
		\label{fig:Learning curve}
        \vspace{-1.5em}
		
\end{figure}

Figure~\ref{fig:Learning curve} shows the reward curves for the training process of the DRL4Route-AC and DRL4Route-GAE algorithms when $\overline{R}$ is set to 20, where the x-axis indicates the number of training iteration rounds and the y-axis indicates the cumulative reward for each episode. The results show that as the number of iteration rounds increases, so does the cumulative reward, reflecting the effectiveness of our DRL4Route framework. Furthermore, the DRL4Route-GAE ultimately performs better than the DRL4Route-AC. It means that the Generalized Advantage Estimation works well to create a trade-off between the bias and variance in Actor-Critic models and thus achieve a more optimal policy.

\subsubsection{Effective of Network Configuration.}
\begin{figure}[hbtp]%hbtp
		\centering
		\includegraphics[width=1 \columnwidth]{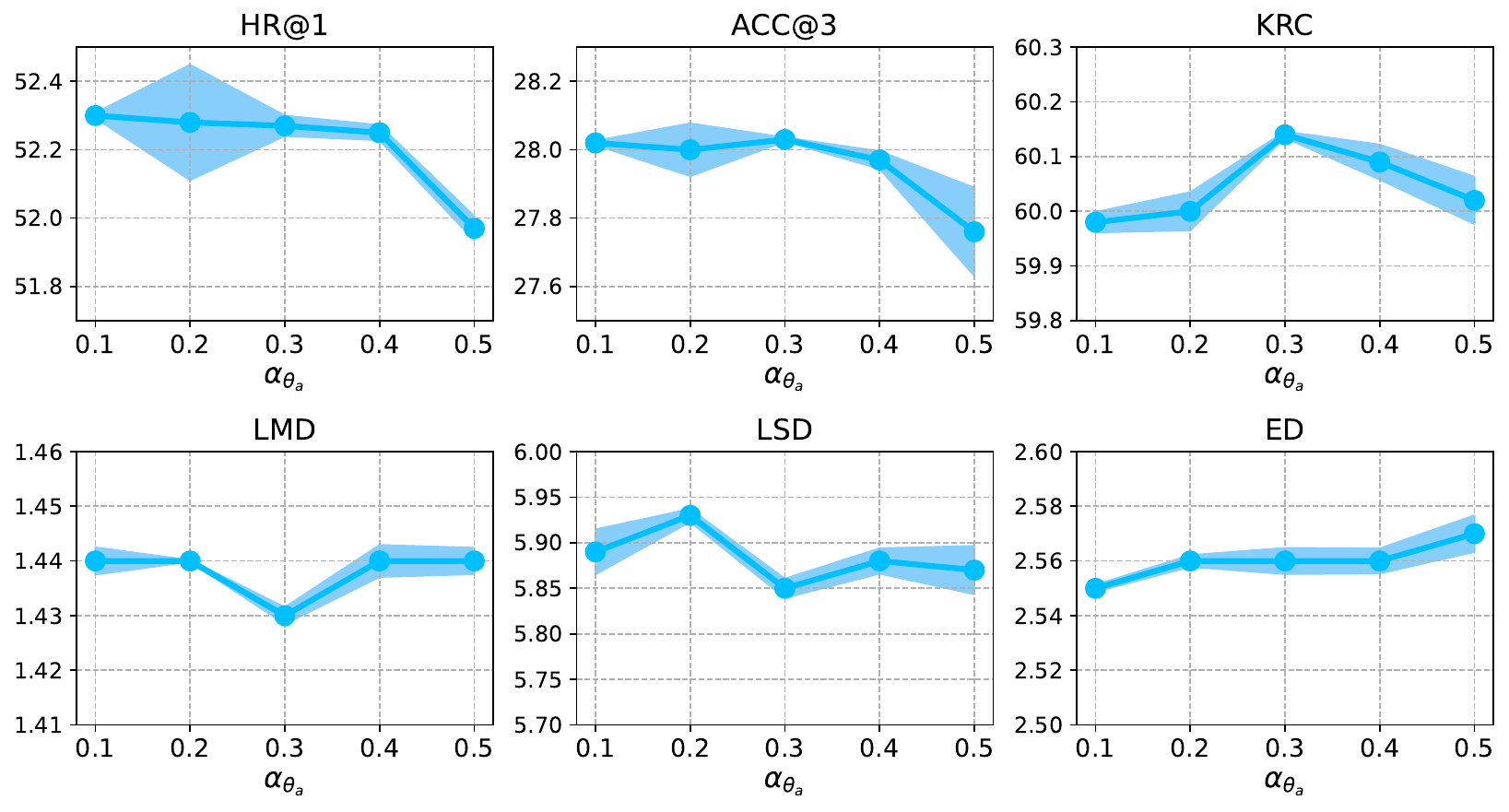}
		\captionsetup{font={small}}
		\caption{Results obtained by setting different $\alpha_{\theta_{a}}$.}
		\label{fig:hyper-parameter}
		
\end{figure}

% We further investigate the influences of hyper-parameter setting, as shown in Figure~\ref{fig:hyper-parameter}.
We further investigate the influences of hyper-parameter setting, as shown in Figure~\ref{fig:hyper-parameter}.
Figure~\ref{fig:hyper-parameter} shows the influences of $\alpha_{\theta_{a}}$ on different metrics while maintaining $\alpha_{c} = 0.1$ and $\alpha_{CE} +\alpha_{\theta_{a}} = 1$ in the Logistics-SH (length 0-25) dataset. From which we have the following observations: (1) Different choices of $\alpha_{\theta_{a}}$ all improve the performance of the algorithm; (2) When the $\alpha_{\theta_{a}}$ is 0.3, the algorithm performs best on the LSD, LMD and KRC metrics, which is a relatively good choice.
\par
\begin{figure}[hbtp]%hbtp
		\centering
		\includegraphics[width=1 \columnwidth]{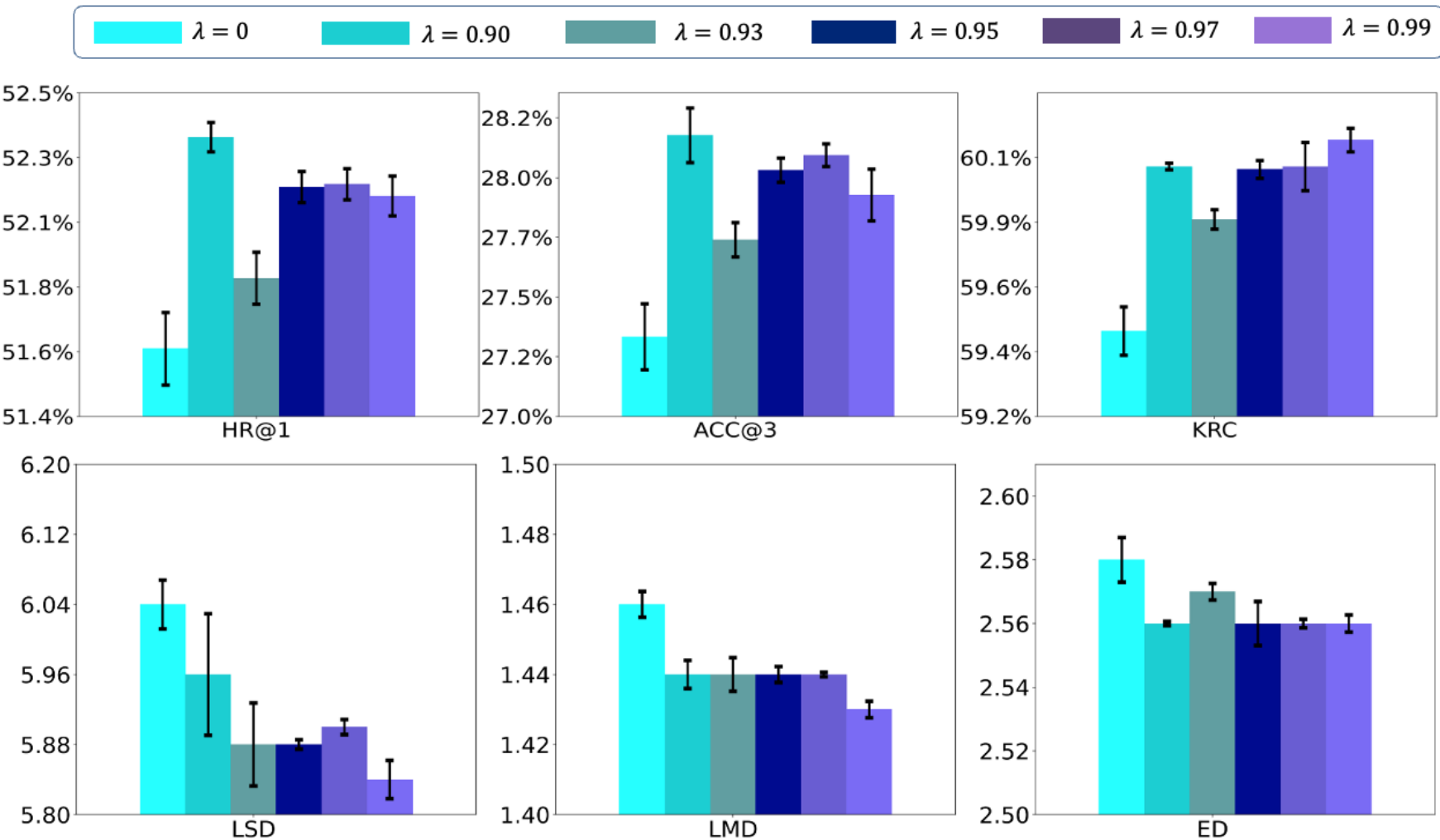}
		\captionsetup{font={small}}
		\caption{Results obtained by setting different $\lambda$.}
		\label{fig:hyper-parameter-lambda}
		
\end{figure}

 We also observe the results of changing $\lambda$ in DRL4Route-GAE and the results are shown in Figure~\ref{fig:hyper-parameter-lambda}. The results show that DRL4Route-GAE performs well on all test metrics for $\lambda$ values between 0.90 and 0.99. When $\lambda$ is 0, DRL4Route-GAE degrades to the DRL4Route-AC algorithm and performs relatively poor on all metrics.


\subsection{Online A/B Test}
DRL4Route-GAE is incorporated into the package pick-up route and time inference system deployed in Cainiao, which supports the main needs of both the user and the courier. The system operates in an environment with hundreds of thousands of queries per day and provides minute-level arrival time estimation for couriers and users, respectively. We leverage DRL4Route-GAE to predict the couriers' decision on the packages to pick up, and the predicted results are used for arrival time estimation. We send a text message to the customer before the courier arrives so the user can be prepared in advance. From the results of the A/B test, DRL4Route-GAE achieves 6.1\% improvements in LSD and 7.5\% improvements in ACC@3 in route prediction. Thus the predicted route deviates less from the courier's actual route, leading to a 6.2\% improvement in Mean Absolute Error at time prediction. Consequently, the customers' complaints because of inaccurate predictions are decreased.
\section{Related Work}

\subsection{Pick-up and Delivery Route Prediction}
\par Pick-up and delivery services are undergoing explosive development, by providing huge convenience to people's daily lives. Massive workers' behavior data are accumulated in the service process, which forms the data foundation of many data mining and learning tasks related to the pick-up and delivery service. For example, \cite{ruan2022service} studies the service time prediction, and \cite{ruan2022discovering} solves the delivery location discovering from couriers' trajectories. By mining workers' behaviors, many efforts are made to improve the service platform \cite{chen2020crowdexpress, liu2018foodnet, guo2018learning, zeng2019last}, such as CrowdExpress \cite{chen2020crowdexpress} and FoodNet\cite{liu2018foodnet}.  And recent years have witnessed increasing attention on pick-up and delivery route prediction from academia and industry. Current works on the topic follow the same research line that trains a deep neural network in a supervised manner, to learn worker's routing patterns from their historical data. For example, DeepRoute \cite{wen2021package} utilizes a transformer encoder to encode unfinished tasks,  and an attention-based decoder to compute the future route. FDNET \cite{gao2021deep} uses a pointer-like \cite{vinyals2015pointer} network for route prediction in food delivery service.  And Graph2Route \cite{wen2022graph2route} uses a dynamic graph-based encoder and personalized route decoder for the PDRP task.
%\par \red{@wen}

\subsection{Deep Reinforement Learning for Seq2seq Models}
\par The sequence-to-sequence (seq2seq) models are the current common models for solving sequential problems, such as machine translation, text summarization, and image captioning. Such seq2seq models suffer from two common problems exposure bias and inconsistency between train/test measurements \cite{8801910}. 
To solve these problems, some DRL methods have been incorporated into the current state-of-the-art seq2seq models.
Policy gradient algorithms were first used in the seq2seq models \cite{xu2015show,wu2018learning,narayan2018ranking}, but only one sample was used for training at each time step, and the reward was only observed at the end of the sentence, resulting in high model variance. Therefore, the baseline reward was designed into the PG model in the following algorithms.
Self-Critic (SC) algorithms \cite{rennie2017self,10.5555/3304222.3304389,hughes2019generating} propose to use the model output from greedy search as the baseline for policy gradient calculation.
Actor-Critic algorithms \cite{liu2017improved, keneshloo2019deep,chen2018fast} propose to train a critic network to output the baseline reward.

However, these RL models suffer from inefficient sampling and high variance. Therefore, Reward Augmented Maximum Likelihood (RAML) \cite{tan2018connecting} proposes to add a reward-aware perturbation to traditional Maximum Likelihood Estimation (MLE), and Softmax Policy Gradient (SPG) \cite{ding2017cold} uses reward allocation to efficiently sample the policy gradient.

\section{Conclusion}
\par  Due to the non-differentiable characteristics of test criteria, deep learning models for the PDRP task fall short of introducing the test criteria into the training process under the supervised training paradigm. Thus causing a mismatch of training and test criteria, which significantly restricts their performance in practical systems. To address the above problem,  we generalize the popular reinforcement learning to PDRP task for the first time, and propose a novel framework called  DRL4Route. It couples the behavior-learning ability of current deep neural networks and the ability  of reinforcement learning to optimize non-differentiable objectives. Moreover, DRL4Route can serve as a  plug-and-play component to boost the existing deep learning models. Based on the framework, we implemented an RL-based model DRL4Route-GAE to solve the route prediction problem in the logistic field. It follows the Actor-Critic architecture, which is equipped with a Generalized Advantage Estimator that can balance the bias and variance of the policy gradient estimates. We empirically show that  DRL4Route achieves significant improvements over the most competitive baseline on the real-world dataset. A direction of future work is to extend  DRL4Route to PDRP tasks in others scenarios, such as food delivery.
\begin{acks}
		\par This work was supported by the National Natural Science Foundation of China (Grant No. 62272033) and CCF-Alibaba Innovative Research Fund For Young Scholars.
	\end{acks}

\bibliographystyle{ACM-Reference-Format}
\balance
\bibliography{reference.bib}

% \vfill\eject
% [23] Reference number 23
% \balance

% \thispagestyle{empty}

\end{document}